\documentclass[review]{elsarticle}
\usepackage{hyperref}

\journal{Journal of \LaTeX\ Templates}
\usepackage{amsmath}
\usepackage{booktabs}
\usepackage{algorithm}
\usepackage{algpseudocode}
\usepackage{graphicx}
\usepackage{subfig}
\usepackage{multirow}
\usepackage{color}
\begin{document}

\begin{frontmatter}
\title{DeepGoal: Learning to Drive with driving intention from Human Control Demonstration}
\author{Huifang Ma\fnref{ZJU}}
\author[mymainaddress]{Yue Wang\fnref{ZJU}\corref{mycorrespondingauthor}}
\ead{wangyue@iipc.zju.edu.cn }
\cortext[mycorrespondingauthor]{Corresponding author}
\author[mymainaddress]{Rong Xiong\fnref{ZJU}\corref{mycorrespondingauthor}}
\ead{rxiong@zju.edu.cn }
\address[mymainaddress]{38 Zheda Road,Zhejiang 310027, China}
\author{Sarath Kodagoda\fnref{UTS}}
\author{Li Tang\fnref{ZJU}}
\fntext[ZJU]{State Key Laboratory of Industrial Control Technology and Institute of Cyber-Systems and Control, Zhejiang University, Zhejiang, China}
\fntext[UTS]{Centre for automotive Systems, The University of Technology, Sydney, Australia}

\begin{abstract}
Recent research on automotive driving developed an efficient end-to-end learning mode that directly maps visual input to control commands. However, it models distinct driving variations in a single network, which increases learning complexity and is less adaptive for modular integration. In this paper, we re-investigate human's driving style and propose to learn an intermediate driving intention region to relax difficulties in end-to-end approach. The intention region follows both road structure in image and direction towards goal in public route planner, which addresses visual variations only and figures out where to go without conventional precise localization. Then the learned visual intention is projected on vehicle local coordinate and fused with reliable obstacle perception to render a navigation score map widely used for motion planning. The core of the proposed system is a weakly-supervised cGAN-LSTM model trained to learn driving intention from human demonstration. The adversarial loss learns from limited demonstration data with one local planned route and enables reasoning of multi-modal behavior with diverse routes while testing. Comprehensive experiments are conducted with real-world datasets. Results show the proposed paradigm can produce more consistent motion commands with human demonstration, and indicates better reliability and robustness to environment change. 
Our code is available at \url{https://github.com/HuifangZJU/visual-navigation}.
\end{abstract}


\end{frontmatter}

\section{Introduction}
In recent automotive driving research, deep learning tries a revolutionizing way for vehicle control, which directly maps raw pixels from camera image to steering commands in an end-to-end manner\cite{Bojarski2016End}\cite{Codevilla2017End}\cite{Xu2017End}. End-to-end(end2end) approach seeks to avoid steps of building an explicit environment model in conventional approach, which include mapping, localization, planned routening and motion planning, etc. Instead, it optimizes driving variations for perception, planning and reasoning in a single network to maximize overall control performance\cite{Bojarski2016End}. In contrast to conventional approach, data for training end2end networks can be collected with relative ease way, i.e., driving around and recording human demonstration control.

\begin{figure*}[!h]
\centering
\includegraphics[width=0.8\textwidth]{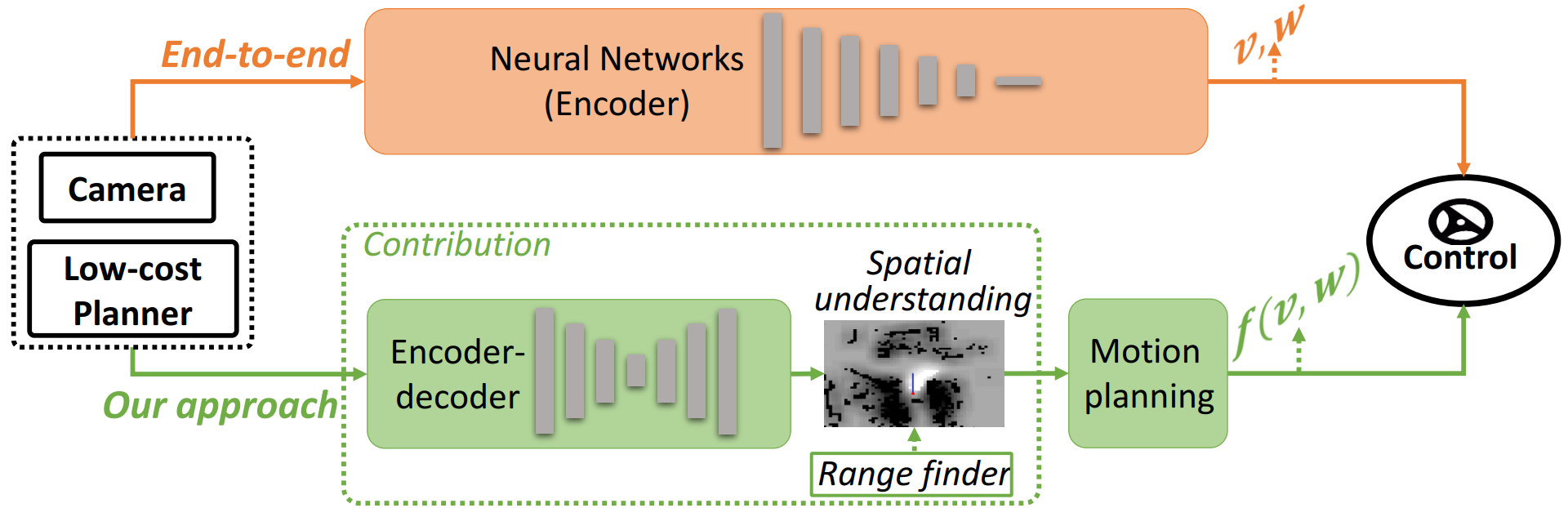}
\caption{Comparison of proposed paradigm with end-to-end model for automotive driving.}
\label{motivation}
\end{figure*}

However, end2end approach faces with the problem of learning very complex mapping in a single network, which needs intensive supervision to handle huge driving variations. Moreover, it prevents intermediate fusion of visual information with other range finder sensors that help much to avoid obstacles. This raises security concerns, as a small error in vision can induce severe consequence for driving in high frequency control loop. Despite these drawbacks, current research often focuses on end2end setting because it allows to look into plenty of challenges. Some recent methods incorporate additional route planner as learning input\cite{gao2017intention}\cite{hecker2018end}\cite{sauer2018conditional}, such as a routed map or a directional instruction, as shown in the top half part of Fig. \ref{motivation}. The planned route captures longer-term motion rules and helps to choose a correct direction upon reaching a fork. It is beneficial and brings performance promotion. Yet the network still lacks transparency of how the planner acts on various driving variations.

In order to address the challenges in end2end approach, we focus on learning an explainable representation 
following the manner of how human drives with route planner. Humans may rely on the planned route in public softwares to figure a direction towards goal, then use visual cues like road semantics to reason where to drive. With the goal-directed area in mind, they perform flexible vehicle control in relating to different driving scenarios. The specific control rules may change, e.g., to follow a lane in urban road nets or to mind unexpected obstacles in campuses. However, a goal-directed visual region is always formed based on the local road situation, which keeps an overall sense of driving direction for vehicle control. We denote this region as driving intention. 

We think this longer-term driving intention region is effective and informative to improve end2end approach, which addresses only visual variations and solves where to go without conventional precise localization and mapping. 
Therefore, we define the pipeline as shown in the bottom half part of Fig. \ref{motivation}. An encoder-decoder structure is adopted to learn the aforementioned driving intention from image perception and a route planner. We follow the work in \cite{hecker2018end} and resort to the planned route in public available navigation softwares from GPS localization. The driving intention region is then projected onto a local navigation score map with range finder data fused to increase reliability. Such a navigation score map encoded with goal-directed information can be directly used for next motion generation, which is also able to explicitly consider more specified motion variations. 


To avoid manual definition and annotation of driving intention on image, we devote to learning from human demonstration. Specifically, human control a vehicle to move and follow a planned route towards goal. Then for each image perception, its traversed area in the near ground satisfies current driving intention and can be projected on the image plane as supervision. The challenge lies in that the demonstration driving only covers a single direction for each fork and cross, while routes planned to different directions can be provided during test. The driving intention is valid only if it holds rationality in regards to the pair of visual observation and local planned route. Thus, we consider the learning task is not pixel-level imitation but structural reasoning, and develop a weakly-supervised model of cGAN-LSTM network utilizing adversarial loss function. The network learns to generate a `fake' driving intention region which is hard to be distinguished from the `real' region maneuvered by human. Then the generated result is punished as a whole to implicitly learn road semantics, planning intention as well as their correlation. Besides, time continuity is considered with a LSTM unit for performance enhancement. The outline of our method is provided in Fig. \ref{outline}. 

In the experiment, a straightforward motion generation method is implemented in a DWA(Dynamic Window Approach)\cite{fox1997dynamic} manner for comparison with end2end system. The proposed pipeline is validated through real-world datasets including previously unseen scenarios to demonstrate generalization performance. Experiments show our approach achieves better performance than state-of-the-art end2end approach and demonstrates improvements in reliability and robustness. To summarize, our main contributions are twofold:
\begin{itemize}
\item An innovate learning-based automotive driving system is developed. The system learns from low-cost GPS-level route planner and images to achieve goal-directed driving intention without precise localization. It eases problem complexity for end2end models and can be efficiently integrated for modular motion planning.

\item A weakly supervised and adversarial learning method is developed through learning from demonstration, the core of which is a cGAN-LSTM network trained with limited single-modal demonstration data. The model is enhanced with time continuity and can be generalized to achieve multi-modal behavior when facing new scenarios.
\end{itemize}

The remainder of the paper is organized as follow: Section \ref{related work} reviews the related works on learning-based approaches for automotive driving. Section \ref{method} illustrates details of the proposed system architecture. Section \ref{experimental result} presents the experimental results, and Section \ref{conclusion} draws a conclusion.

\section{Related Work}\label{related work}
Conventional pipeline of vehicle control includes
mapping, localization, path planning and motion planning. The result of localization, i.e. a reference path or a goal in vehicle local coordinate, and the result of path planning, i.e. a grid map with(out) semantics, are fed into motion planing to generate final control command. The system is thus sensitive to environment change and calls for a lot of work to improve performance on separate modules. In this section, we give a brief review of two learning-based systems aimed at improving the conventional automotive driving, each following a different approach to the system design: the end2end approach and the intermediate recognition approach.
\paragraph{End-to-end approaches} Recently, end2end method learned from human demonstration becomes popular in automotive driving. The intrinsic merit is that the performance of intermediate stages in conventional system architecture may not be aligned with the ultimate goal, namely, the control of the vehicle. With this idea, \cite{Bojarski2016End} firstly proved powerful ability of CNN to steer a vehicle directly from vision input. Codevilla, et,al.\cite{Codevilla2017End} then proposed to learn the driving model to compute motion command via conditional imitation learning, which incorporates high-level command input to consider the repeatability of imitation learning. The work in \cite{gao2017intention} collected control commands from existing local planner (Dynamic Window Approach\cite{fox1997dynamic}) and proposed a two-stage approach to relax prior knowledge for localization. This relies on the path-planning results, as the form of navigation is to learn expected motion commands using a residual neural network. The work in \cite{hecker2018end} adopted 360-degree surround-view cameras along with planned routes information from commercial maps to learn an end2end driving model. Their work has utilized GPS signals as well as public map to generate steering angles and speeds based on a RNN. As reported in their evaluation, it has unavoidably incorporates human intervention. The work in \cite{amini2018variational} proposed to estimate a variational network to get a full probability distribution over the possible control commands; however, when combined with specific navigation indicators, they still solve an accurate form of certain control command. Compared with this category, our work relies on similar input while achieves intermediate representation in robot local coordinate without relying on precise geometric transformations. We consider the problem space of end2end control learning is more complicated than ours, as the motion states of the vehicle are coupled with the visual understanding.
\paragraph{Intermediate Recognition approaches}
Another group of works focus on
CNN based recognition tasks to facilitate automotive driving. One category is following the scene parsing problem to recognize common traffic elements in an image, such as vehicles, pedestrians and cyclists detection \cite{redmon2018yolov3,liu2016ssd,chen2018deeplab,luo20173d,li2018scale,yang2016exploit}. Another idea goes a step further towards vehicle direct usage, including regression of distance to other cars\cite{chen2015deepdriving}, recognition of traversable region\cite{barnes2017find}\cite{tang2017one}, or road attributes prediction\cite{sauer2018conditional}, etc. The work in \cite{chen2015deepdriving} intended to find an intermediate traffic semantics by regressing distances to the lanes as well as other road users.Tang et al, \cite{tang2017one} applies mapping techniques to achieve weakly supervised learning of traversable regions. Wang et al, \cite{wang2017scalable} proposed an on-line learning mechanism to deal with the appearance change of traversable region without referring to the massive data. In \cite{seff2016learning}, a model to infer road layout and vehicle relative pose is developed given imagery from on-board cameras, utilizing the public Google Street View and OpenStreetMap. The work in \cite{sauer2018conditional} proposed intermediate affordances to facilitate driving, including both vehicle relative pose to the road and recognition of traffic signs, while the human-defined affordances may not be adaptive to different driving scenarios.
These works have eventually yielded an intermediate representation for vehicle control, while may still needs further reasoning for motion control. And some of the learned perception result still rely on accurate pose estimation from conventional approach.

\section{Methods}\label{method}
\begin{figure*}[!h]
	\centering
	\includegraphics[width=\textwidth]{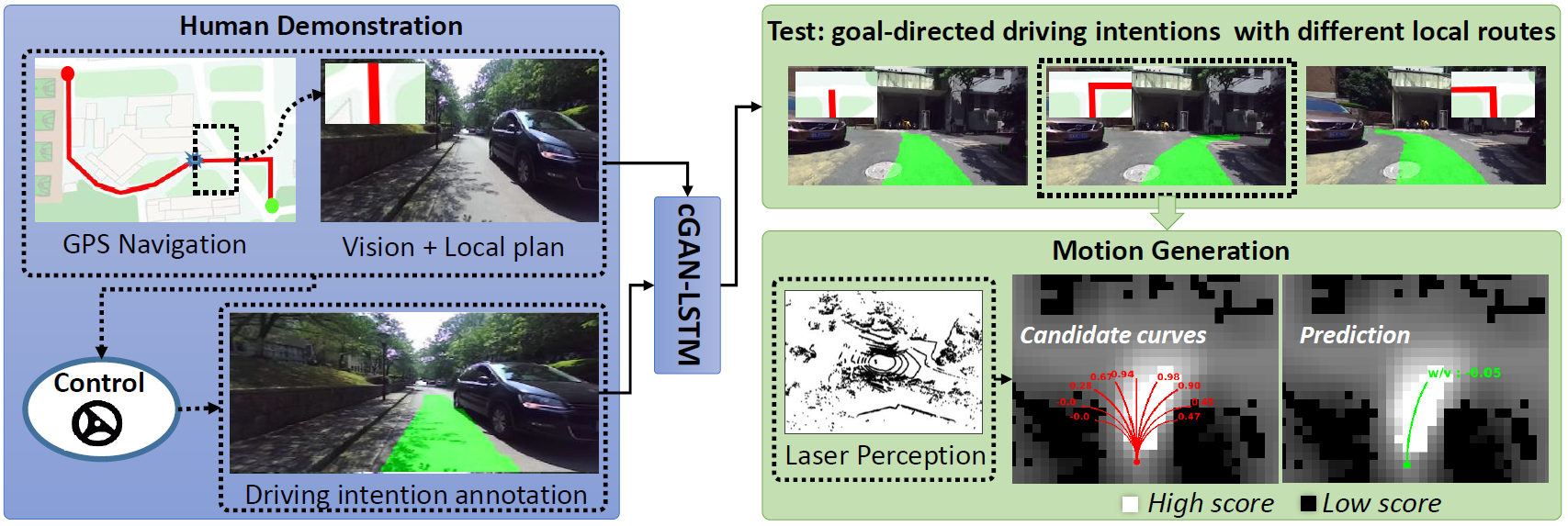}
	\caption{Outline of our approach. A cGAN-LSTM model is utilized to learn driving intention from human demonstration with local planned route. While testing, the model generates corresponding driving intentions following both local routes and road structures. The driving intention is then integrated with concurrently collected laser data and rendered into a navigation score map. Based on this, we generate motion by scoring candidate driving curves.}
	\label{outline}
\end{figure*}
This section introduces the proposed driving paradigm in detail. The system architecture is shown in Fig. \ref{outline}, blue box shows the procedure of learning from human demonstration and green boxes show the test application. The core of the system is a cGAN-LSTM network, which takes front-view image and local planned route from public navigation software as input. The model learns to generate a goal-directed driving intention region on the road area to indicate supposed future control. When testing in strange scenarios with different local planned route, the model is able to generate corresponding driving intentions towards different directions. For motion generation, the driving intention is then projected on vehicle near ground and integrated with concurrent laser perception to render a navigation score map. Finally, we implement a straightforward method to generate control command by scoring candidate driving curves. The specific illustrations of each step is provided in the following parts.

\subsection{Weakly-Supervised driving intention learning}
\subsubsection{Network design}
\begin{figure*}[!ht]
	\centering
	\includegraphics[width=\textwidth]{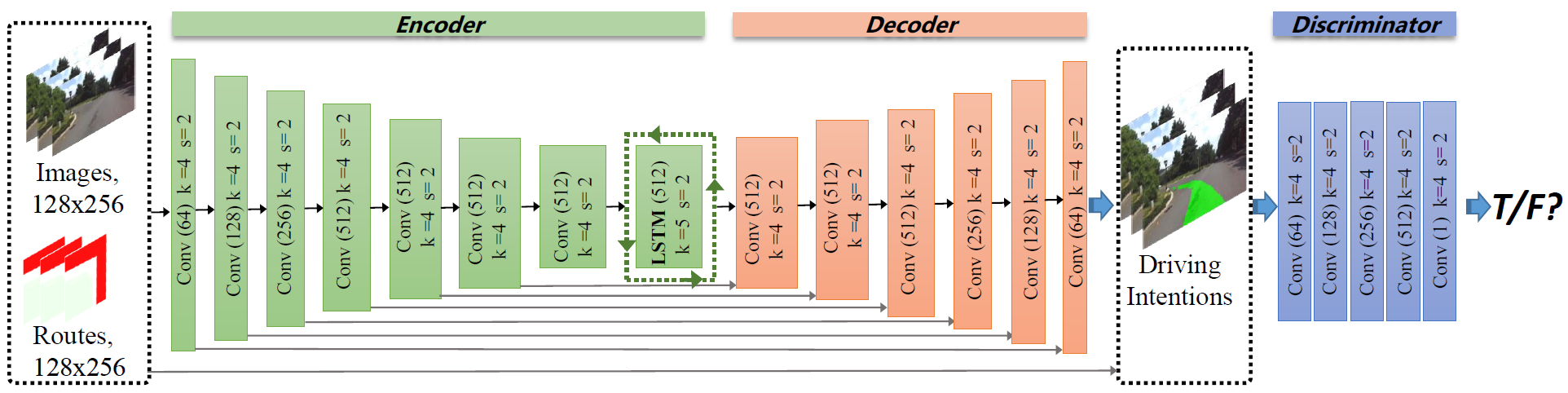}
	\caption{Model architecture of cGAN-LSTM. Front-view images combined with local planned routes are fed into a UNet structure to generate goal-directed driving intentions. The middle of the UNet is inserted with a LSTM-unit to incorporate time continuity. The predictions are then concatenated with input images to go through the discriminator.}
	\label{model}
\end{figure*}

We frame the intention learning as a structure reasoning process to follow both road situation in image and planning intention in local planned route. Since human demonstration only covers a single direction for each fork and cross while routes planned to different goals can be provided during test, we do not treat the learning as a pixel-level imitation and regression. Instead, we adopt the recent GAN\cite{goodfellow2014generative} model which utilizes adversarial loss to generate and evaluate the network output as a whole. GANs consist of a generator and a discriminator. The fake output from generator is trained to be similar with the real data so as to cheat the discriminator. Thus, the generator eventually learns to produce an overall reasonable result following the distribution in provided dataset. We implement a network structure of cGAN-LSTM specifically for driving intention generation, the model structure is provided in Fig. \ref{model}.

The network structure has referred to the work in \cite{isola2017image}, which follows the design of conditional GAN\cite{mirza2014conditional} and utilizes a UNet\cite{ronneberger2015u} as generator. As implied, UNet is an encoder-decoder structure with skip connections to keep lower-stage features. We use less layers in our case since the generation task does not need to recover complete textures of the image. Our model treats both the image and the planned route as prior conditions. The two inputs together with the generated driving intention are fed into the discriminator for evaluation. Since the intentions are continuous in both time and space domain during driving, a LSTM(Long Short-term Memory) unit is inserted after the last encoder layer to capture series relation, which at the same time guarantees minimum parameter increase.


Let's consider $k-1$ steps of the former visual perception, the sequential input images are denoted as $I_{[t-k+1,t]}$ and the corresponding local planned routes are denoted as $R_{[t-k+1,t]}$, for each time $t$. We expect to learn a visual driving intention towards goal, denoted as $V_t$, at current image. Thus, the problem is formulated as $G : \{I_{[t-k+1,t]}, R_{[t-k+1,t]}\}\rightarrow V_t$. Since previous approaches have found it beneficial to mix the GAN objective with a more traditional loss, such as L1 distance\cite{isola2017image}, the objective function is the sum of two weighted loss for the considered $k$ steps:
\begin{equation}
L = arg\,min_{G}\,max_D\,\sum_{t}^{t-k+1}L_{cGAN}(G,D) + \lambda L_{L1}(G)
\end{equation}
where $\lambda$ is the weight parameter.

The first item is the standard cGAN objective function:
\begin{equation}
\begin{aligned}
L_{cGAN}(G,D) = &E_{I_t,R_t,V_t}[logD(I_t,R_t,V_t)] + \\
&E_{I_t,R_t}[log(1-D(I_t,R_t,G(I_t,R_t)))]
\end{aligned}
\end{equation}
and the second item is a patch-wise $L1$ distance from generated intention to the provided real driving intention:
\begin{equation}
L_{L1}(G) = E_{I_t, R_t, V_t}[||V_t - G(I_t,R_t)||_1]
\end{equation}

The cGAN-LSTM model ensures generated intention to consider both current road structure in image perception and different intentions in local planned routes. It has implicitly learned their inherent correlation by adversarial training and can be generalized to allow for different driving intentions when confronting new scenarios.

\subsubsection{Data preparation}\label{labelgeneration}
To achieve the learning of driving intention, we need to provide data of image perception, local planned route as well as annotation of driving intention region. More importantly, their correlation needs to be specifically established. This part illustrates how we make the data ready for training network and performing comprehensive experimental evaluation.

\begin{figure*}[!ht]
	\centering
	\includegraphics[width=\textwidth]{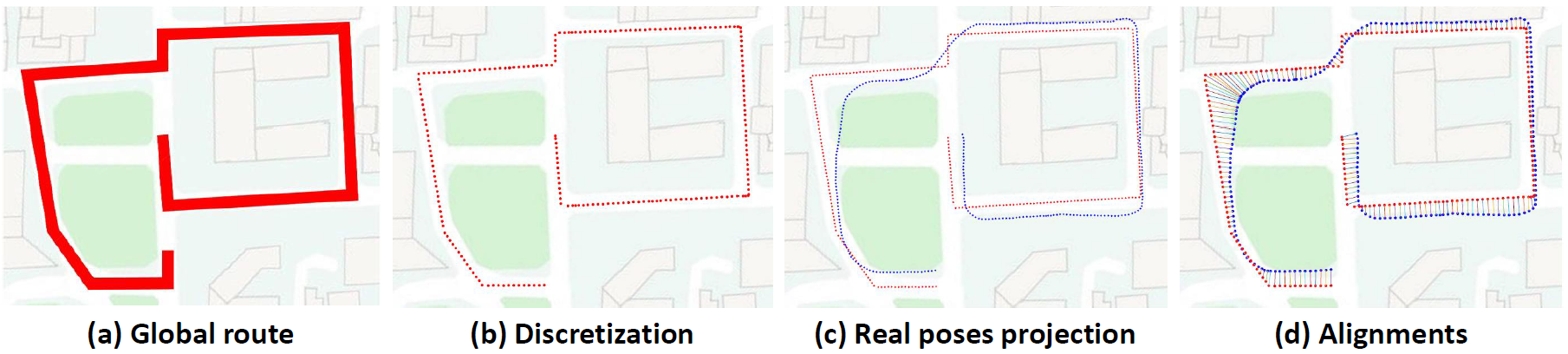}
	\caption{Procedure to get off-line local planned routes.}
	\label{map rendering}
\end{figure*}

\textbf{Local planned route} 
We devote to follow human's manner which uses public navigation softwares to get local planned route with GPS signal. Thus one possibility is to enable the communication between vehicle to such a APP during demonstration. Nevertheless, the interface to real-time synchronized view of local planned route is not make public for research usage. Besides, the model performance under different GPS localization errors needs to be carefully considered and experimentally evaluated. This may lead to substantial workload for on-line data collection and rendering. Therefore, we develop an off-line route rendering method, which makes use of the spatial alignment between public map and vehicle demonstrated trajectories. The procedure is shown in Fig. \ref{map rendering}.

Given public map data from Baidu Map\footnote{https://map.baidu.com/}, the global route $R$ is annotated in a similar manner to that the navigation software uses, as shown with the red line in Fig. \ref{map rendering}(a). Then, the route $R$ is discretized to route points $R_d$ in Fig. \ref{map rendering}(b).  After human demonstration driving along the global route, we obtain the vehicle poses as shown with blue dots in Fig. \ref{map rendering}(c), denoted as $T_r$.

The spatial alignment from vehicle pose to the routed public map is now the task of aligning two sets of planar points $R_d$ and $T_r$. Here, we use the DTW (dynamic time warping)\cite{berndt1994using} algorithm which is commonly adopted in the time domain to warping time series data:

\begin{equation}
DTW(T_r,R_d) = min\,\frac{1}{K}\sqrt{\sum_{k=1}^{K}w_k}
\end{equation}
where $K$ is the warpping length, and $w_k=(i,j)_k$ is the warping weight between  $T_r$ to $R_d$.

The step-by-step optimization objective is:
\begin{equation}
\begin{aligned}
\gamma(i,j) = w(T_r(i),R_d(j)) + min\{\gamma(i-1,j-1),\gamma(i-1,j),\gamma(i,j-1)\}
\end{aligned}
\end{equation}
where $\gamma$ is the accumulated series distance. Specifically, a geometric warping criterion is adopted in our scenario, and $w$ is defined as the euclidean distance from projected vehicle pose to the center route point:
\begin{equation}
w(T_r(i),R_d(j)) = ||(T_r(i)),R_d(j)||_2
\end{equation}

The aligned result is shown in Fig. \ref{map rendering}(d). By assigning the heading direction of each road, we can crop desired local planned route under various experiment settings. The vehicle pose used for data preparation is calculated with the conventional localization approach in our previous work\cite{tang2017one}. 

\begin{figure*}[!ht]
	\centering
	\includegraphics[width=0.8\textwidth]{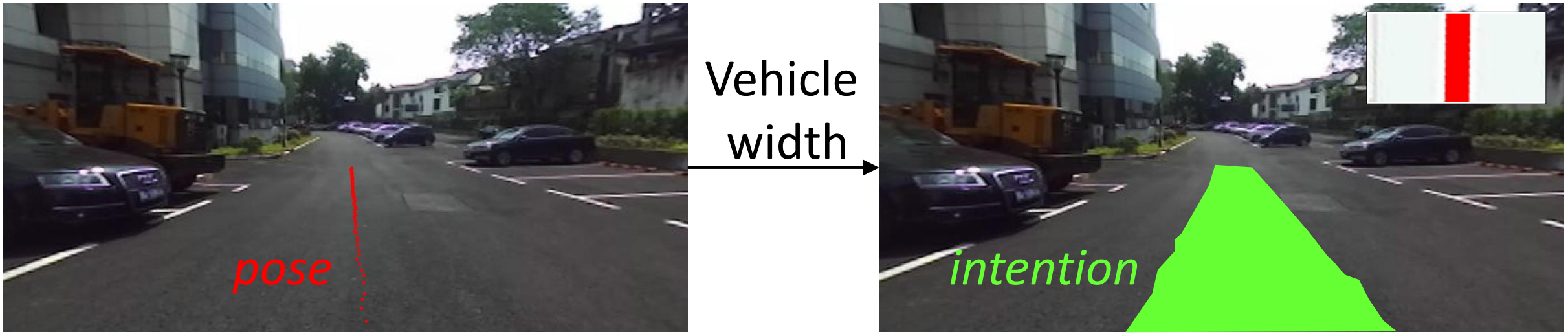}
	\caption{Driving intention annotation. Left: vehicle projected poses; Right: annotated driving intention region.}
	\label{annotation}
\end{figure*}

\textbf{Driving intention annotation.} We annotate driving intention region from human demonstration, which projects vehicle traversed area under specified local planned route on current image, as shown in Fig. \ref{annotation}.

Human control a vehicle to move and follow a specified route towards goal. For each image, vehicle future poses in the near ground are first projected on the image plane. Then the poses are dilated with vehicle width to indicated current driving intention. The annotation certainly satisfies planning intention in the planned route and road semantics in image. This idea is similar to the drivable region annotation work in \cite{barnes2017find}. However, they do not differentiate region directions and further do not learn its relation with the planned route. The other region on the image are then similarly labeled with \textit{obstacle} and \textit{unknown}\cite{barnes2017find} utilizing the projection of concurrently collected laser perception.

\subsubsection{Training}
We consider four time steps to train the cGAN-LSTM network: 0.9s in the past, 0.6s in the past, 0.3s in the past, and the current frame, similar to the experiment setting in \cite{hecker2018end}. As the straight road sections are much longer than the turning sections, the straight perceptions are down-sampled to around one sixth to keep a same quantity with that of turning perceptions.

For parameter optimization, we firstly train a basic-model of cGAN without LSTM unit, following the common procedure of one gradient descent on $D$ and then one step on $G$. The basic model network is trained with stochastic gradient descent(SGD) at a learning rate of 0.0002, and momentum parameters of $\beta_1 = 0.5$, $\beta_2 = 0.999$. We train basic model to 200 epochs with a batch size of 12. Then cGAN-LSTM model is fine-tuned based on the pre-trained parameters of basic model, and the encoder part of UNet is fixed to keep a stability of the network. The cGAN-LSTM model is trained additionally around 20 epochs. At inference time, the generator net runs in exactly the same manner as during the training phase.

\subsection{Motion generation with driving intention}\label{spatial understanding}
\begin{figure*}[!ht]
	\centering
	\includegraphics[width=\textwidth]{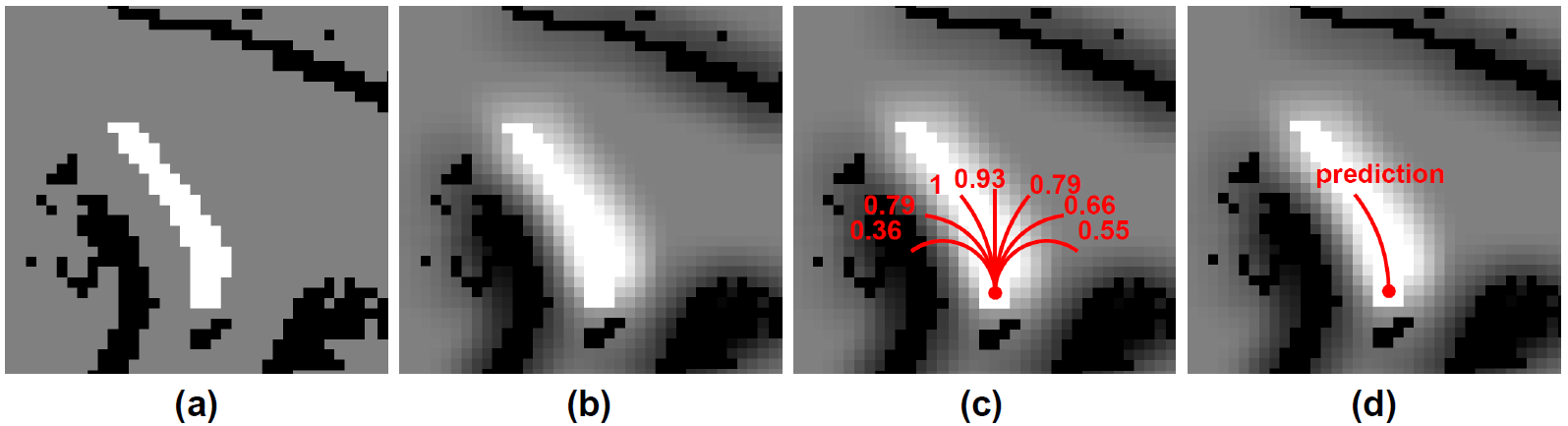}
	\caption{Motion generation with learned driving intention. (a) Projection of visual driving intention with laser perception integrated; (b) Navigation score map modeled with Gaussian kernel; (c) Candidate driving curves with their score labeled; (d) Final control command generated with the driving intention.}
	\label{space2control}
\end{figure*}

The learned driving intention preserves a learned region which is highly adaptive to fuse with other sensors and motion variations to ensure vehicle safety.  Considering vehicle usually runs on smooth roads, we make an assumption that the road area in the near front of vehicle can be modeled with a flat plane. Thus, driving intention learned to follow road structure in image plane can be projected on robot local coordinate with camera calibration parameters. Then local driving projection is then integrated with concurrent laser perception to render a navigation cost map, based on which, a motion generation method is implemented by scoring candidate driving curves. The procedure is shown in Fig. \ref{space2control}.

The white grids(0.5m$\times$0.5m) in Fig. \ref{space2control}(a) shows projected driving intention in vehicle local ground. The black grids indicate the obstacle perception for concurrent laser data. To consider neighboring influence, driving intention grids are assigned with positive Gaussian kernels and obstacle grids are assigned with negative Gaussian kernels, which forms the navigation score map used for motion generation, as shown in Fig. \ref{space2control}(b).

In order to keep the task tractable, we chose to generate motion in an DWA manner, i.e., to produce candidate driving commands and estimate the best one. Following the work in \cite{Bojarski2016End}\cite{amini2018variational}, we present the steering command as driving curvature, denoted as $\frac{1}{r}$, where $r$ is the turning radius in meters. Since the vehicle constantly adjust its control based on visual perception, it is reasonable to assume that vehicle keep a uniform motion during a short time clip. Resultantly, vehicle future trajectory can be modeled with a curve in the local navigation score map. 

Specifically, the $90^{\circ}$ space ahead of vehicle is divided into different number of direction sections in relating to different requirement for control precision. We then generate a same number of candidate driving curves whose curvatures range evenly form $[-0.2, 0.2]$, as shown in Fig \ref{space2control}(c) which is an example of generating seven candidate curves. Each curve can be estimated with a score based on the navigation score map. Then the final command is the curve with the highest score, indicating direction towards goal, as implied in Fig. \ref{space2control}(d). The motion generation is straightforward while efficient to show its usage and compare with end2end approach, the incorporation of more specified motion variations is the future work.

\section{Experiments}\label{experimental result}
This section reports the experiment results of the proposed approach, including experiment set up, performance of driving intention generation, and performance of motion generation compared with the state-of-the-art end2end approach.

\subsection{Data sets}
\begin{figure}[!h]
\centering
\includegraphics[width=0.85\textwidth]{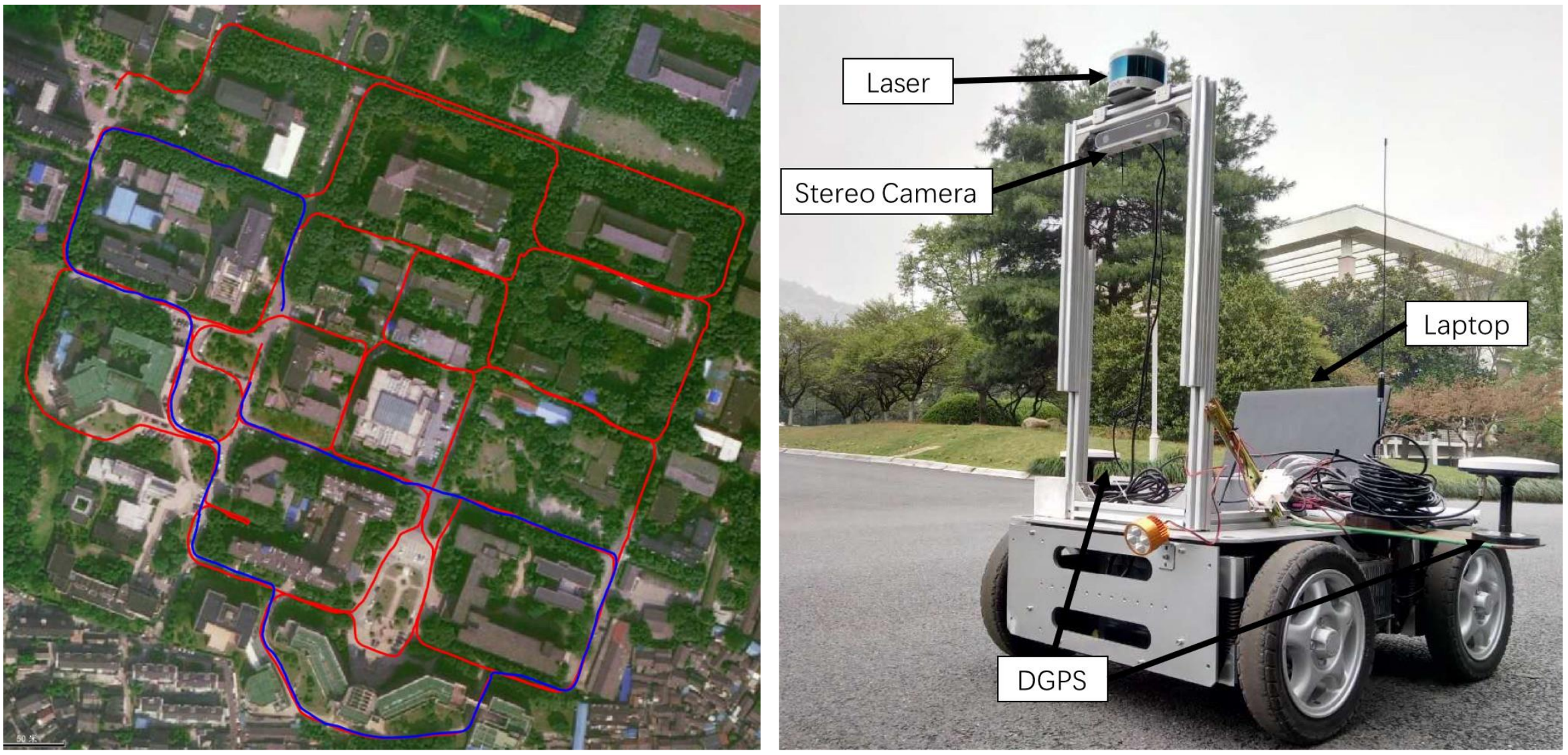}
\caption{Data collection route and experiment vehicle.}
\label{hardware}
\end{figure}

Experiment data is collected with a real vehicle running in our campus, which has been extensively adopted in conventional automotive driving research\cite{ding2018laser,tang2019topological,yin2018locnet}. The data collection route is shown in the left of Fig.\ref{hardware}. Blue line shows the training route with a length of 1.2km. Red line shows the test route with a total length of 4.8km. The overlap sections of the two routes basically runs in a bi-directional manner. The vehicle used for data collection is shown in the right side of Fig.\ref{hardware}, which is a four-wheeled mobile vehicle equipped with a ZED stereo Camera, a Velodyne VLP-16 laser scanner and a D-GPS. Only images from the left camera of ZED are used with a resolution of 314$\times$648 pixels. The training data involves 21 times of demonstration driving at different time over three days, covering varying weather/illumination conditions. Each demonstration driving contains $\sim$7000 frames of observation. The test data contains $\sim$25000 frames of observation.

\subsection{Intention generation result}
We first present the result of driving intention generation. Since the representation of driving intention region is a novel contribution in our work, we propose two new criteria along with the common used IOU for quantitative evaluation.

\begin{figure}[!h]
\centering
\includegraphics[width=0.5\textwidth]{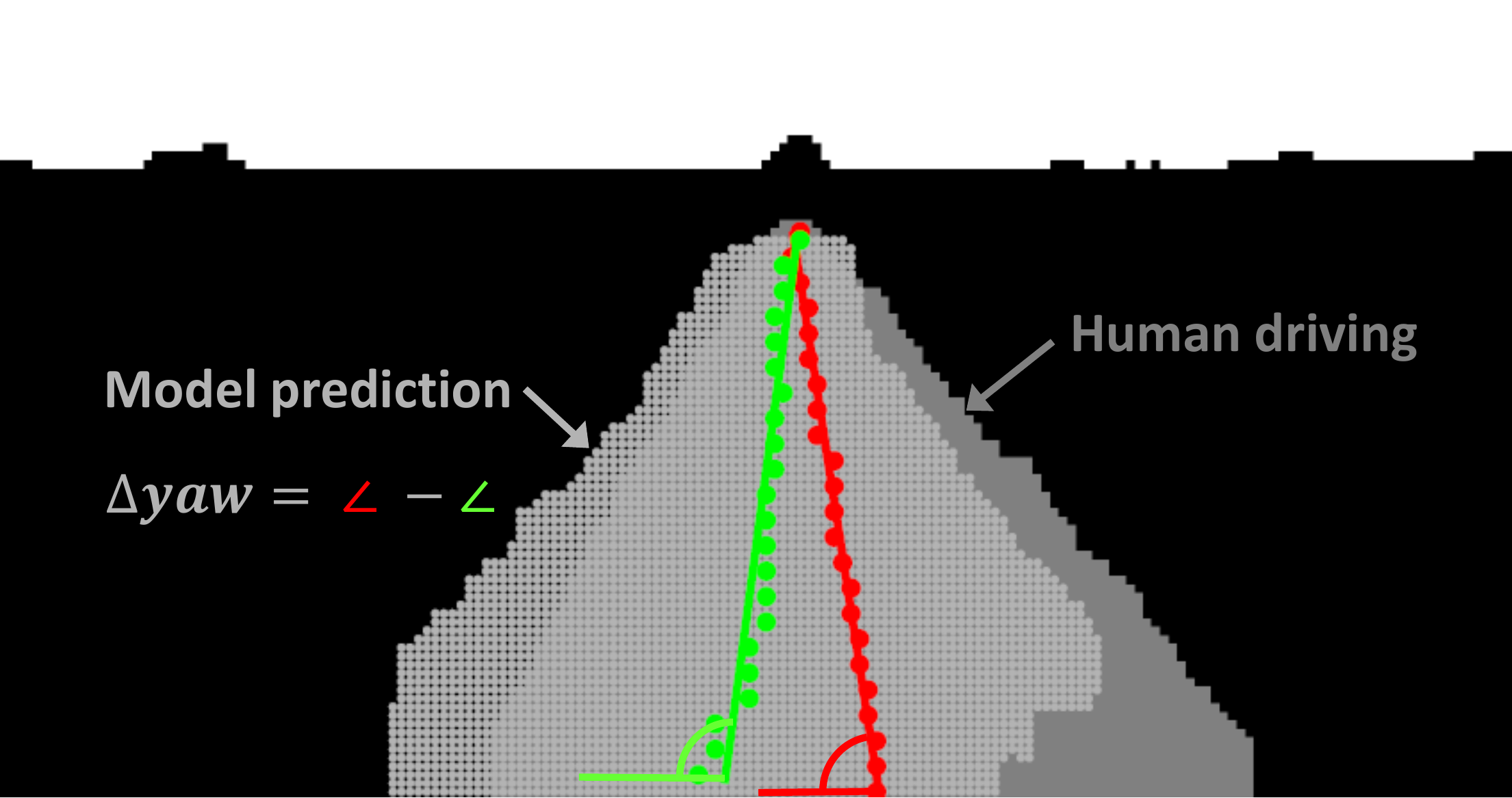}
\caption{Illustration of visual evaluation. The dark gray shows the future control of human demonstration and the light gray grids imply model prediction. Red dots and green dots show the center lines of human driving and model prediction respectively, which have been both downsampled for better visualization.}
\label{visual evaluation}
\end{figure}

\textbf{Evaluation Metrics.}  As shown in Fig. \ref{visual evaluation}, dark gray represents the demonstrated human driving and light gray grids shows the predicted intention. Green dots and red dots show the center line of the two intentions respectively. Based on these representation, we employ three criteria for visual goal-directed intention evaluation. The first is IOU, intersection over union between prediction and human demonstration. The second is $cover\_rate$, ratio of predicted central line falling inside demonstrated intention region, which is measured as the percentage of green dots falling in the dark gray region. And the last is $\Delta{yaw}$, angle difference of driving directions between prediction and human demonstration in image plane. The driving direction is measured as the angle between central lie and horizontal line.

\subsubsection{Network performance}
We test the visual result with two models of $cgan\_basic$ and $cgan\_lstm$. The $cgan\_basic$ model does not include a LSTM unit. The result is presented in Table \ref{path generation}.
\begin{table}[!h]
\caption{P{\scriptsize ERFORMANCE} {\scriptsize OF} D{\scriptsize RIVING}  I{\scriptsize NTENTION}  G{\scriptsize ENERATION}}
\centering
\begin{tabular}{ccccc}
	\toprule[0.025cm]
	Model& iou\%&$cover\_rate\%$&$\Delta{yaw}$\\			
	\cmidrule{1-4}
	\textit{$cgan\_basic$} &62.06& 95.9&13.71\\	
	\cmidrule{1-4}	
	\textit{$cgan\_lstm$}&78.01&98.9 &9.05 \\			
	\bottomrule[0.025cm]
\end{tabular}
\label{path generation}
\end{table}

\begin{figure}[!h]
\centering
\includegraphics[width=\textwidth]{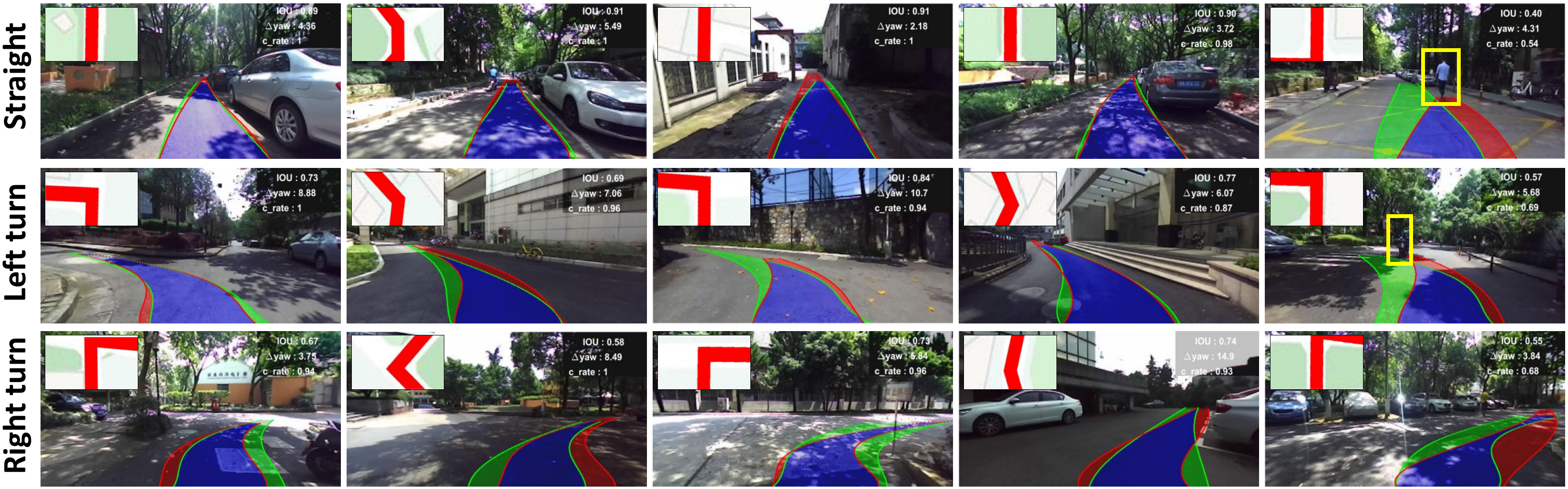}
\caption{Visual results from cGAN-LSTM. Green, red and blue colors represent prediction, human demonstration and their intersection respectively. The local planned routes are shown in the top left corner of the image and the evaluation figures are annotated in the top right corner of the image. The intention boundaries are fitted with splines in these figures.}
\label{result_turn}
\end{figure}

From the table we can see, $cgan\_lstm$ outperforms $cgan\_basic$ for all the three criteria. The LSTM unit has shown its effectiveness to model time continuity and shape the driving intentions. The $cover\_rate$ of both models has exceeded 95\%, which implies the predicted intention basically follows the road structure, as the human demonstration region is certainly on the road area. The model of $cgan\_lstm$ has achieved more improvement on the IOU which is 78.01\%, and the $\Delta{yaw}$ which is around 9 degrees. A higher IOU value implies more similarity to human demonstration. However, we do not seek a one-hundred-percent result, since the driving intention is not defined with an explicit boundary and there can be some pixel-level shape variations. In this situation, $\Delta{yaw}$ shows more importance for direction measurement. We consider 9-degree is a favorable value, since the case in Fig. \ref{visual evaluation} has a 11-degree different in $\Delta{yaw}$, where the predicted intention is structural close to human demonstration. More qualitative results from $cgan\_lstm$ are presented in Fig. \ref{result_turn}.

The first row of Fig. \ref{result_turn} presents some cases for driving on straight roads. The straight road is relatively easy task for intention generation. And the mainly challenge lies on the recognition of obstacles, as shown in the last figure of this row. It provides an example when confronting with a pedestrian which is annotated with a yellow box. The generated intention has turned to a more central direction than that of human demonstration. Thus, it causes lower criteria values.

The last two rows show some cases for left turns and right turns respectively. The proposed model has generated correct driving intentions following various local road situations.  For the turning classes, we see more shape changes to that of human demonstration. The result satisfies the design of adversarial loss, which focus more on the structural rationality rather than pixel-level imitation. Besides, the yellow box in the second row indicates a cyclist where the predicted intention has also avoided. Therefore, the proposed model seems to show an ability of avoiding dynamic obstacles on the road without explicit annotation. 

\subsubsection{Robustness to Localization errors}
We assume to utilize public navigation softwares from GPS signal. As the GPS signal commonly provides rough localization results, this section discusses the model performance under different localization errors. In order to achieve it, we have randomly added horizontal and vertical offsets when rendering local planned route, as illustrated in Section \ref{labelgeneration}. The random offset goes into three levels, easy, moderate and hard, with each level corresponds to a localization error of $0 m\sim1m$, $1 m\sim2.5m$, and $2.5 m\sim5m$. The result is presented in Table \ref{robustness to localization}
\begin{table*}[!h]
\caption{R{\scriptsize OBUSTNESS} {\scriptsize TO} L{\scriptsize OCALIZATION}  E{\scriptsize RRORS}}
\centering
\begin{tabular}{lccccccccc}
	\toprule[0.025cm]
	\multicolumn{1}{l}{\multirow{2}{*}{}} &
	\multicolumn{3}{c}{minor(0m$\sim$1m)}&\multicolumn{3}{c}{moderate(1m$\sim$2.5m)}&\multicolumn{3}{c}{hard(2.5m$\sim$5m)}\\
	\cmidrule(lr){2-4} \cmidrule(lr){5-7} \cmidrule(lr){8-10}
	Model &iou & $c\%$&$\Delta{yaw}$&iou & $c\%$&$\Delta{yaw}$&iou & $c\%$&$\Delta{yaw}$\\	
	\cmidrule{1-10}	
	\textit{$basic$} &61.8&96.2&13.9&61.6&95.8&14.2&61.5&95.9&14.2\\
	\cmidrule{1-10}
	\textit{$lstm$}  &76.5&98.4&9.9&76.1&98.3&10.3&75.9&98.4&10.5\\
	\bottomrule[0.025cm]
	\multicolumn{10}{l}{{\scriptsize abbreviations: basic($cgan\_basic$), lstm($cgan\_lstm$), $c\%$($c\_rate\%$)}}
\end{tabular}
\label{robustness to localization}
\end{table*}

\begin{figure}[!h]
\centering
\includegraphics[width=\textwidth]{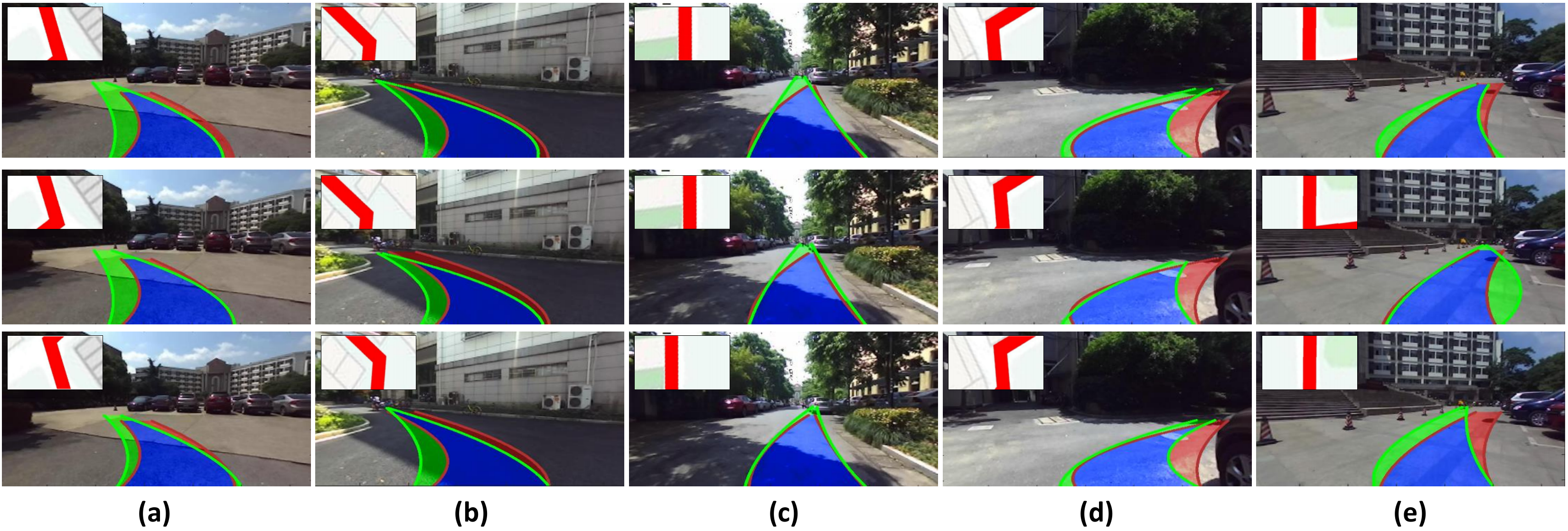}
\caption{Robustness to localization errors. From top to bottom: minor(0m$\sim$1m), moderate(1m$\sim$2.5m) and hard(2.5m$\sim$5m) errors. Green, red and blue colors represent the prediction, human demonstration and their intersection respectively.}
\label{localization}
\end{figure}

As can be seen in the table, the two models have basically achieved a stable performance given different level of route offsets. To compare with the result from center-view routes in Table \ref{path generation}, only $\Delta{yaw}/deg$ has a slight increase with the growing of localization errors, while the other two criteria have remained in a similar value to the previous result. The model has shown robustness to potential localization errors. Moreover, it verifies the proposed pipeline can get a local reference path towards goal without common requirement to precise geometric localization. Some visual result from $cgan\_lstm$ with different level of route offsets are shown in Fig. \ref{localization}. The offset in local planned route does not have a severe influence for intention generation, as the annotated route clearly conveys an overall sense of direction within visible range. 


\subsubsection{Discussion$\colon$ multi-modal behavior}
\begin{figure}[!h]
\centering
\includegraphics[width=0.8\textwidth]{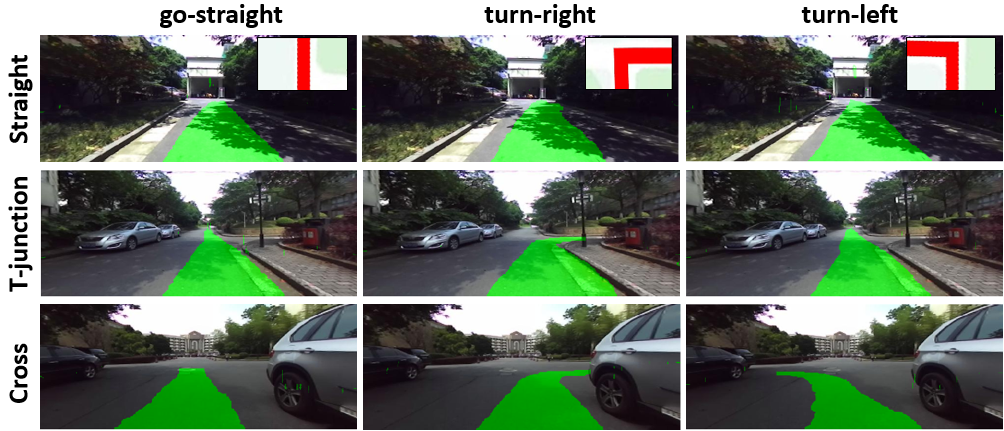}
\caption{Multi-modal driving behaviors with different road types. From top to bottom: one-way straight road, T-junction road, and cross area. Green color shows the original output from the network.}
\label{simulation1}
\end{figure}

Annotation from human demonstration only validates a single driving intention with pre-defined local planned route, while multi-modal behaviors are presented when approaching intersections and open areas. For a further qualitative evaluation, three fake planned routes are made to test the model, which are intuitively viewed as \{go-straight, turn-right, turn-left\}. These planned routes are all used to generate driving intentions on the test image perceptions, and Fig. \ref{simulation1} provides some visual result from a direct network output.

The three rows show the road types of straight, T-junction and cross respectively. The local planned routes are shown in the top right conner on the first row. For the first two classes, there are local planned routes that may not be allowed on current road situation. In this case, the model can still generate reasonable intentions following road structure. While for the last class, where all planned routes can be performed, the model generates driving intentions accordingly. Here, we did not use a spline to fit the intention boundary, which better shows the direction difference on network outputs for different planned routes.

\begin{figure}[!h]
\centering
\includegraphics[width=\textwidth]{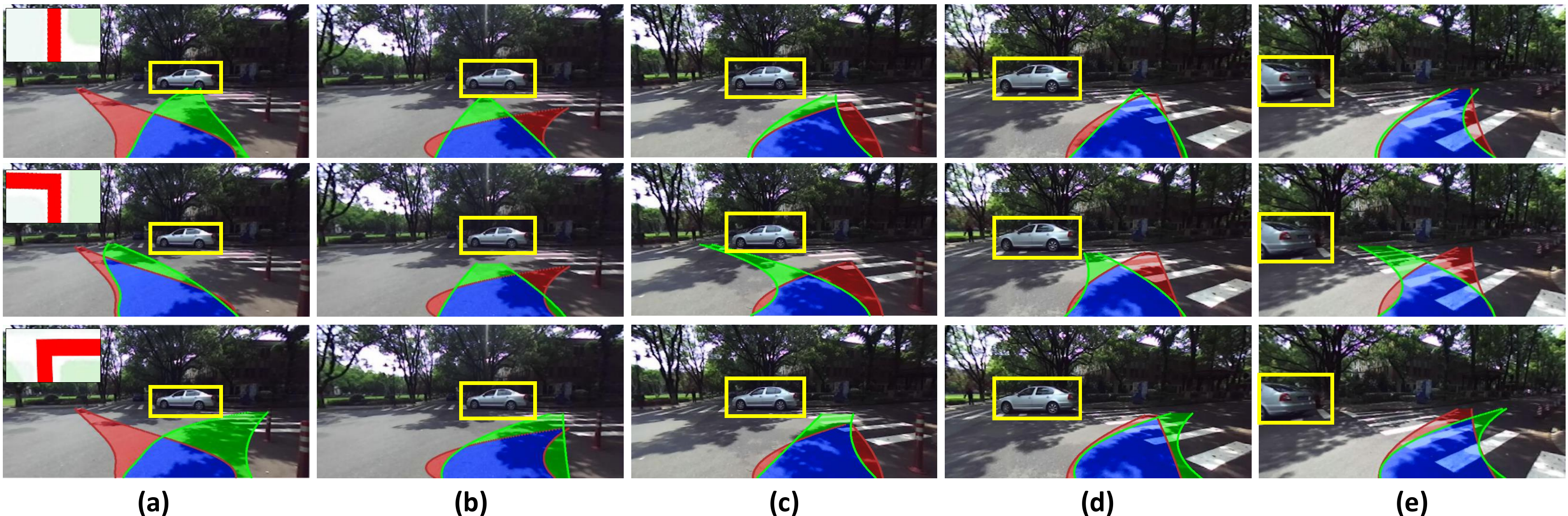}
\caption{Multi-modal driving behaviors when facing a moving car. Green, red and blue show the result from cGAN-LSTM, pix2pix and their intersection respectively. The intention boundaries are fitted with splines in these figures.}
\label{simulation2}
\end{figure}

Fig. \ref{simulation2} has specifically presented a group of images when facing a moving car with different planned routes. The result is visually compared with the model of $pix2pix$\cite{isola2017image} which does not include planned routes for intention generation.  When there are dynamic obstacles appear in front of the road, the proposed model accordingly adjust the intention generation, which further implies the weakly-supervised model has learned to avoid obstacles without explicitly annotation.

\subsection{Motion generation result}
To show the effectiveness of proposed driving intention on motion generation, a straightforward method is performed by scoring candidate driving curves. For comparison to end2end approach, we implement the network structure in \cite{hecker2018end}, which generates direct control commands with public planned route. Model in \cite{hecker2018end} outputs driving commands as velocity and angular speed. We calculate driving curvature as $c_t = \frac{w_t}{c_t}$, for each time $t$ referring to \cite{amini2018variational}. As for the `ground truth' from human demonstration, the actual trajectory curvature is calculated for comparison.

\begin{figure}[!h]
\centering	
\includegraphics[width=0.8\textwidth]{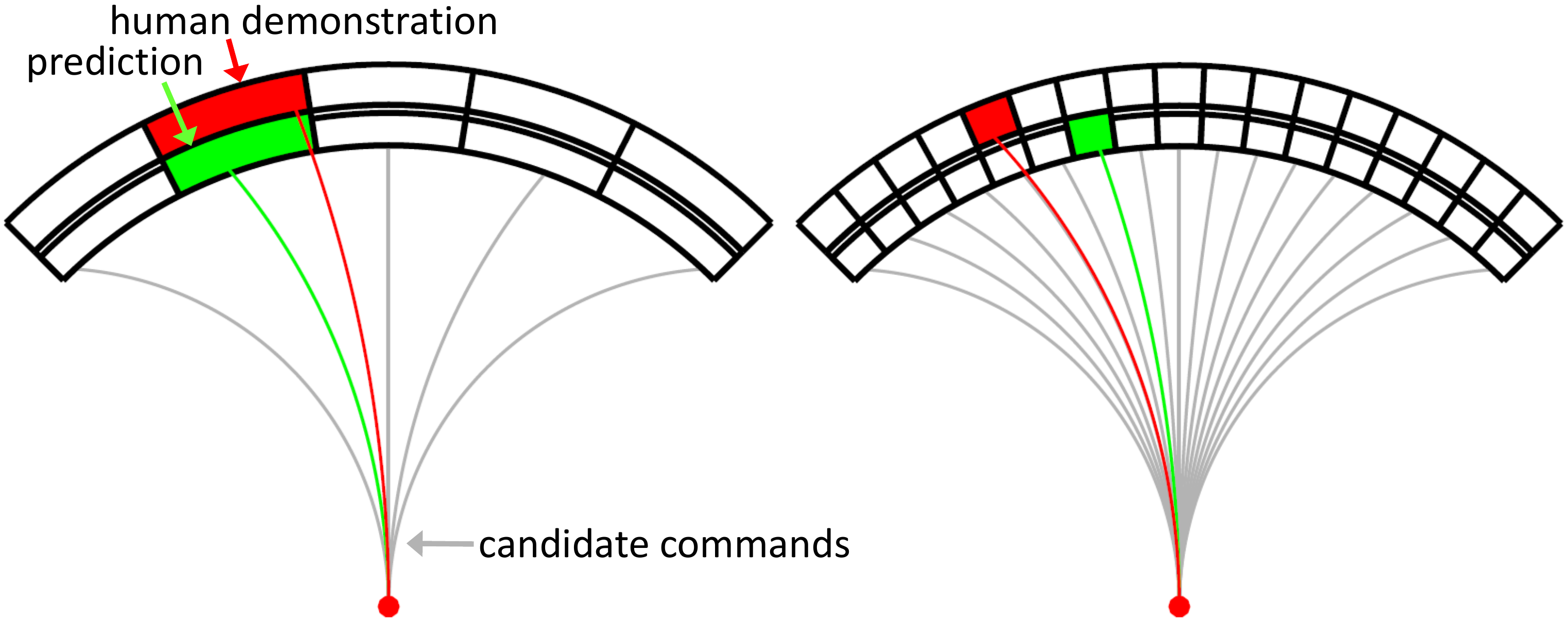}	
\caption{Evaluation criterion for motion generation. Gray curves show the candidate driving commands in relation to the control precision. Green curve indicates the command with a highest score in the navigation score map. And red color shows the ground truth maneuvered by human demonstration.}
\label{planning_evaluation}
\end{figure}

\textbf{Evaluation Metrics.}
To quantitatively evaluate the system performance, we compute the motion prediction accuracy for different control precisions. The definition of true positive is illustrated in Fig. \ref{planning_evaluation}. 

The black dials implies different motion resolutions, for which the proposed method generates a same number of candidate motion commands, as shown with the gray curves. Green grid indicates final model prediction and red grid indicates human demonstration. Then the prediction accuracy is measured with the grid distance $\Delta g$ from model prediction to human demonstration:

\begin{equation*}
accuracy = \frac{\sum_{t=0}^{T} \textbf{[}|pred_{t}-human_{t}|\leq \Delta g\textbf{]}}{T}
\end{equation*}
where $T$ is the total step of test dataset. $[\cdot]$ equals to 1 if the formulation inside is true, otherwise equals to 0. For comprehensive evaluation, we provide quantitative results of three settings of $\Delta g =\{0,1,2\}$ under control resolutions for 3 to 23. The resolution of 3 means the prediction only distinguish the direction from right, left to straight. And a resolution of 23 represents a control precision of less than $4^{\circ}$.
\subsubsection{Model performance}
\begin{figure}[!h]
\centering
\subfloat[Overall accuracy]{\includegraphics[width=0.33\textwidth]{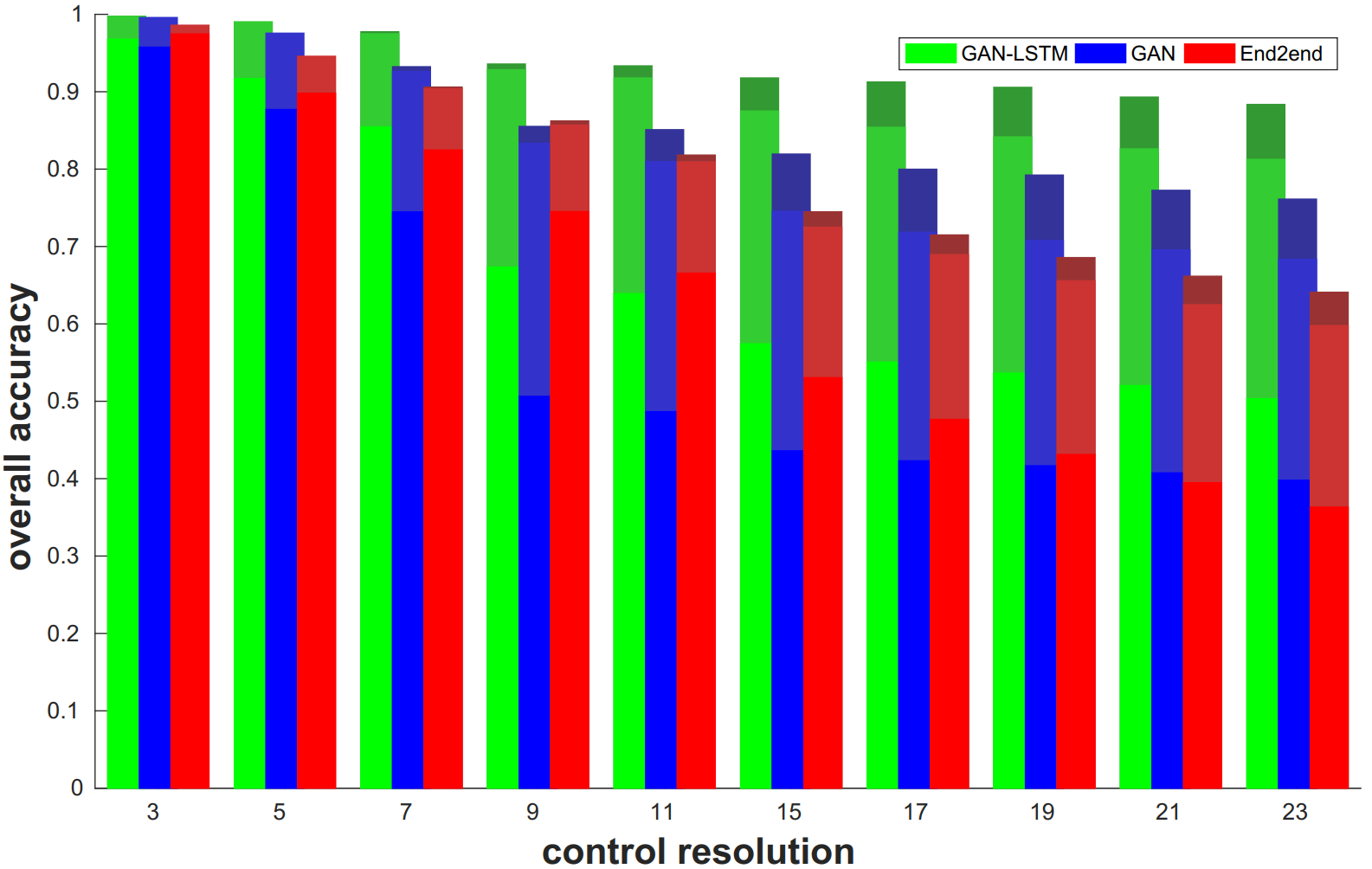}}
\subfloat[Straight road accuracy]{\includegraphics[width=0.33\textwidth]{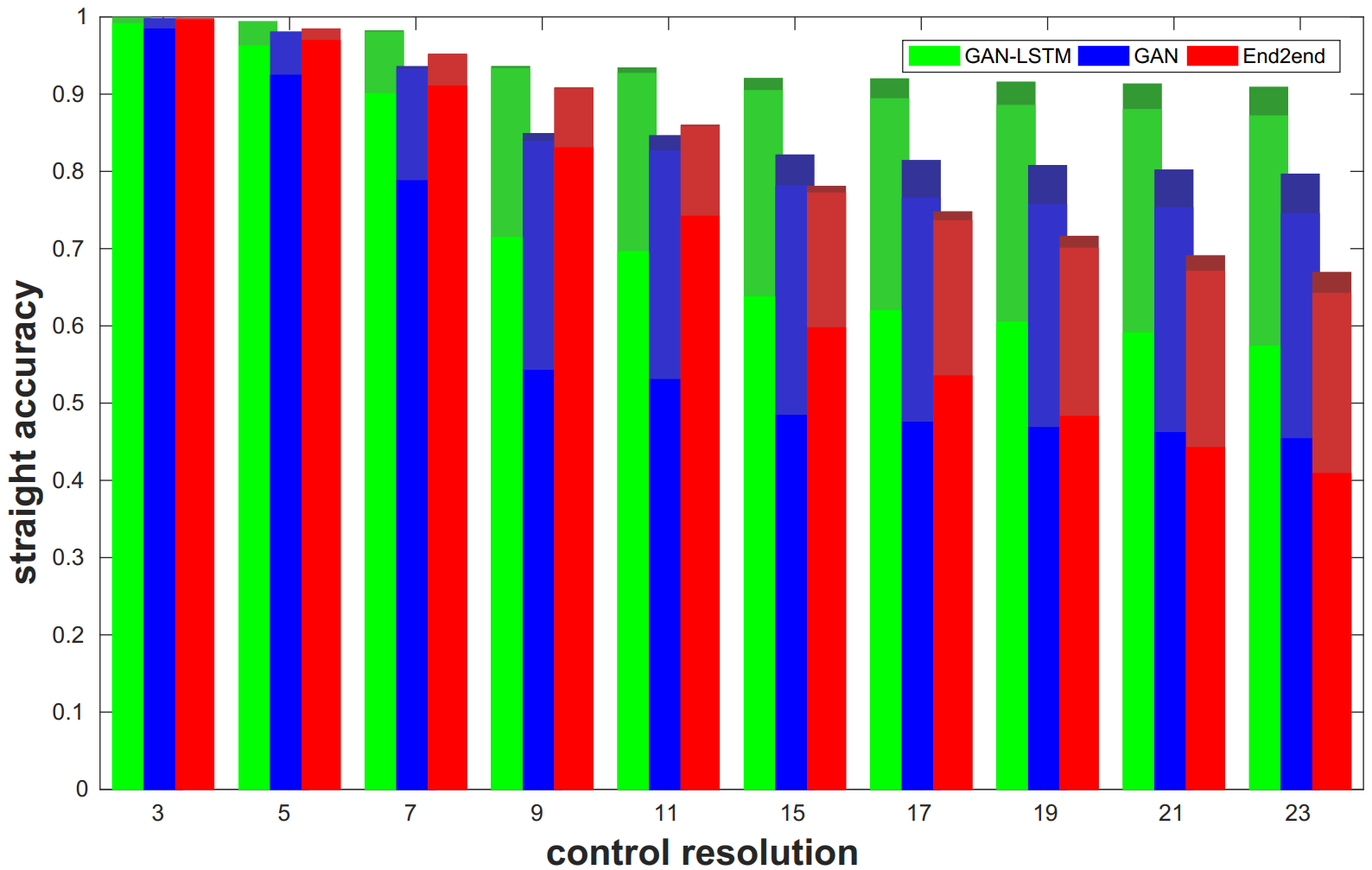}}
\subfloat[Turning roa accuracy]{\includegraphics[width=0.33\textwidth]{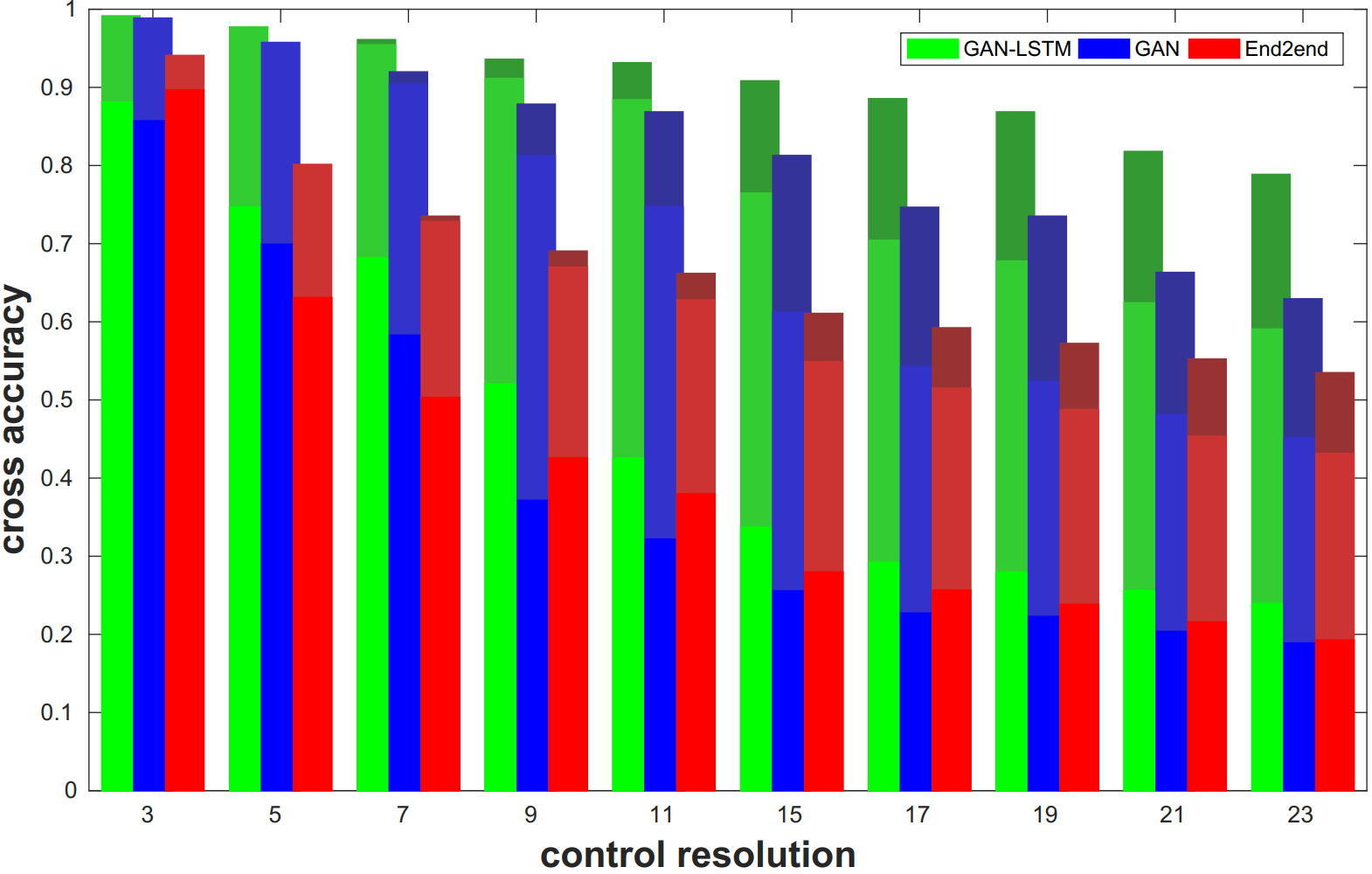}}

\caption{Motion prediction performance. Green, blue, and red represent the performance of $cgan\_lstm$, $cgan\_basic$ and end2end respectively. For each individual model result, color from light to dark represent a grid distance 0,1,2 for $\Delta g$.}
\label{planningaccuracy}
\end{figure}

We have tested our two models of $cgan\_basic$ and $cgan\_lstm$ to compare with end2end\cite{hecker2018end} method. The results with different motion resolutions are shown in Fig. \ref{planningaccuracy}.

The three figures present accuracy on complete test route, long straight test route, and turning route respectively. For an overall accuracy measurement, $cgan\_lstm$ achieves better performance than end2end method under most settings and shows better robustness to different control resolutions. When the control resolution goes up, $cgan\_lstm$ also generates more candidate curves for a careful search to better fit for the driving intention. This shows the benefit to separate driving variations on perception from that on motion in the proposed system. The visual driving intention is highly adaptive to different motion requirements, regardless of specific control precision in the demonstration driving, as the intention region explicitly considers local road situation and leaves a room for a second motion module to follow.

\begin{figure}[!h]
\centering
\subfloat[cGAN-LSTM]{\includegraphics[height=0.45\textwidth]{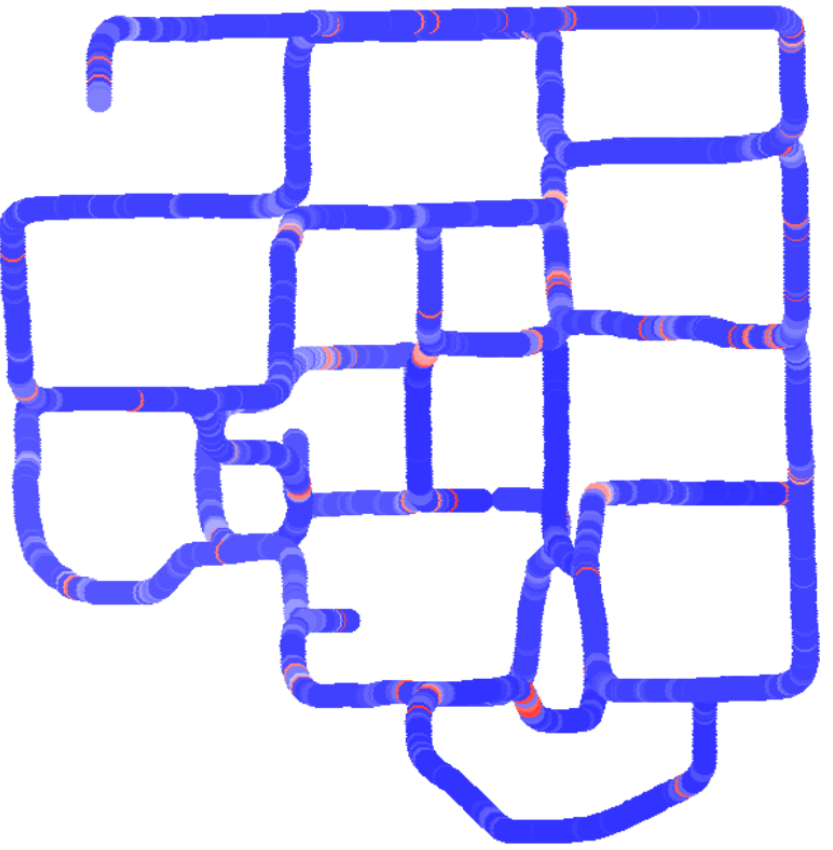}}
\hspace{0.01\textheight}
\subfloat[End-to-end]{\includegraphics[height=0.45\textwidth]{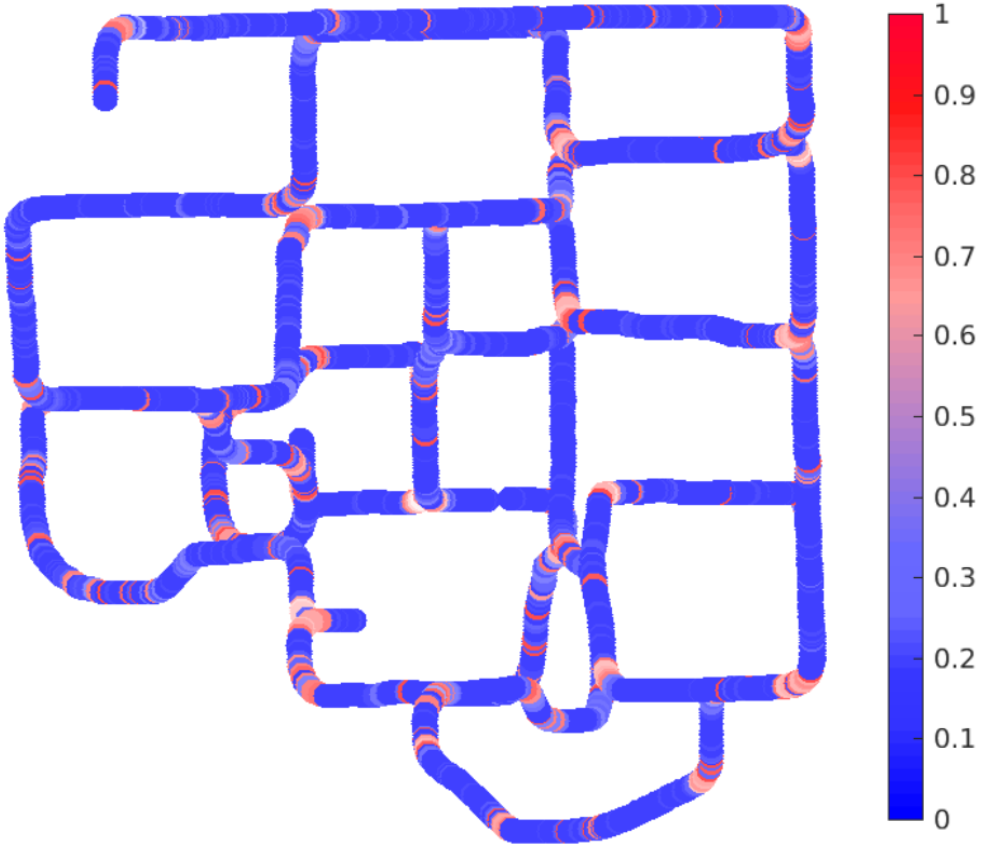}}
\caption{Error rate density along the test route. The results come from a control resolution of 7-grid. Different colors represent different error levels, and the blue the better.}
\label{section}
\end{figure}

In contrast, end2end approach learns a numerical mapping from visual input to output. Thus, end2end outputs a same result for different control resolutions, and shows steady decrease when the evaluation gets more strict. For the performance on separate straight class and turning class, advantage of the proposed model is more significant for the turning road sections which is more challenging due to the complex road situation. A more intuitive comparison of error rates along test route between $cgan\_lstm$ and end2end are provided in Fig. \ref{section}.

\begin{figure*}[!h]
\centering
\includegraphics[width=\textwidth]{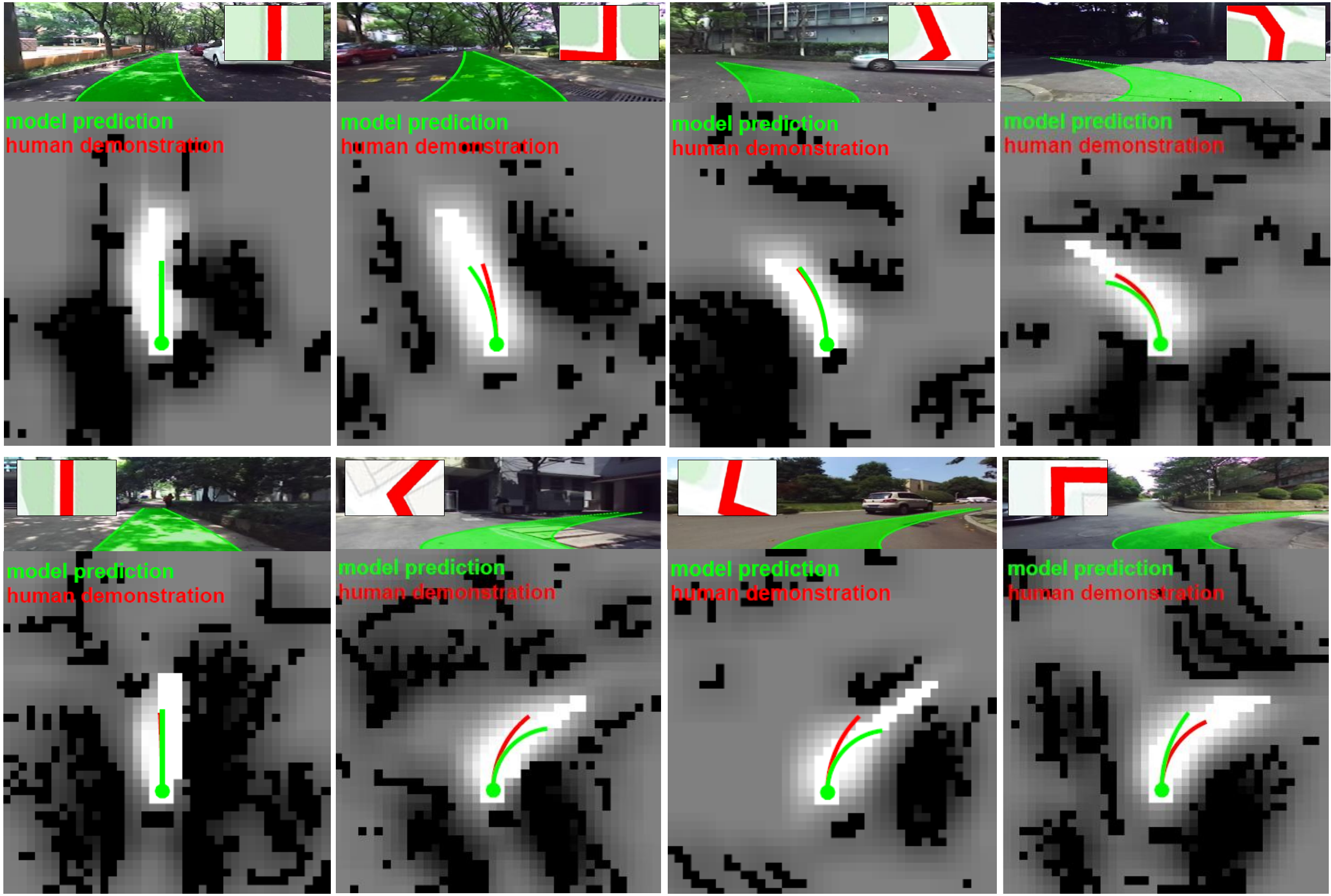}
\caption{Motion generation result compared with human demonstration. For each case, top figure shows the visual result and bottom figure shows the generated motion on navigation score map. Green and red colors represent the model prediction and human demonstration respectively.}
\label{planning}
\end{figure*}

As the figure shows, end2end method have more errors when vehicle approaching turning road sections. We consider the reason may be the numerical difference of demonstration control when facing a same turning class. A big turn requires a small control curvature and a small turn may require a big control curvature. Besides, human may change the driving command during turning for obstacle avoidance. This also increases the intra-class variation for similar perceptions. Thus, it may be difficult for end2end training to consider both variations in perception and planning. In contrast, for the long straight road section, most commands stay near-linear and unchanged for a long time, which can be efficiently learned by both models. Some visual result from the proposed system is shown in Fig. \ref{planning}.

%

\subsubsection{Discussion$\colon$ robustness to time delay}
\begin{figure}[!h]
	\centering
	\includegraphics[width=0.9\textwidth]{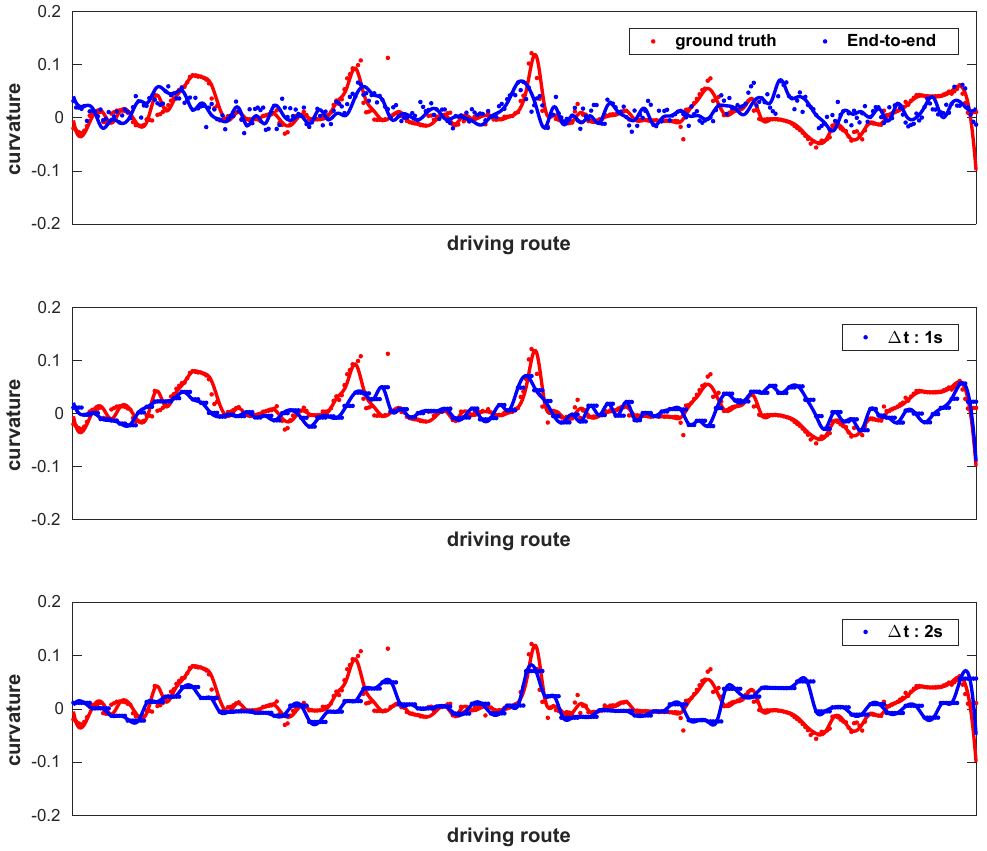}
	\caption{Comparison of end2end method with human demonstration when visual prediction is delayed. From top to bottom: no time delay, 1s delay, and 2s delay. Blue color indicates the model prediction and red color denotes human demonstration. The dots denoted actual data points, which are fitted to a smooth line to show the prediction tendency.}
	\label{delay1}
\end{figure}

\begin{figure}[!h]
\centering
\includegraphics[width=0.9\textwidth]{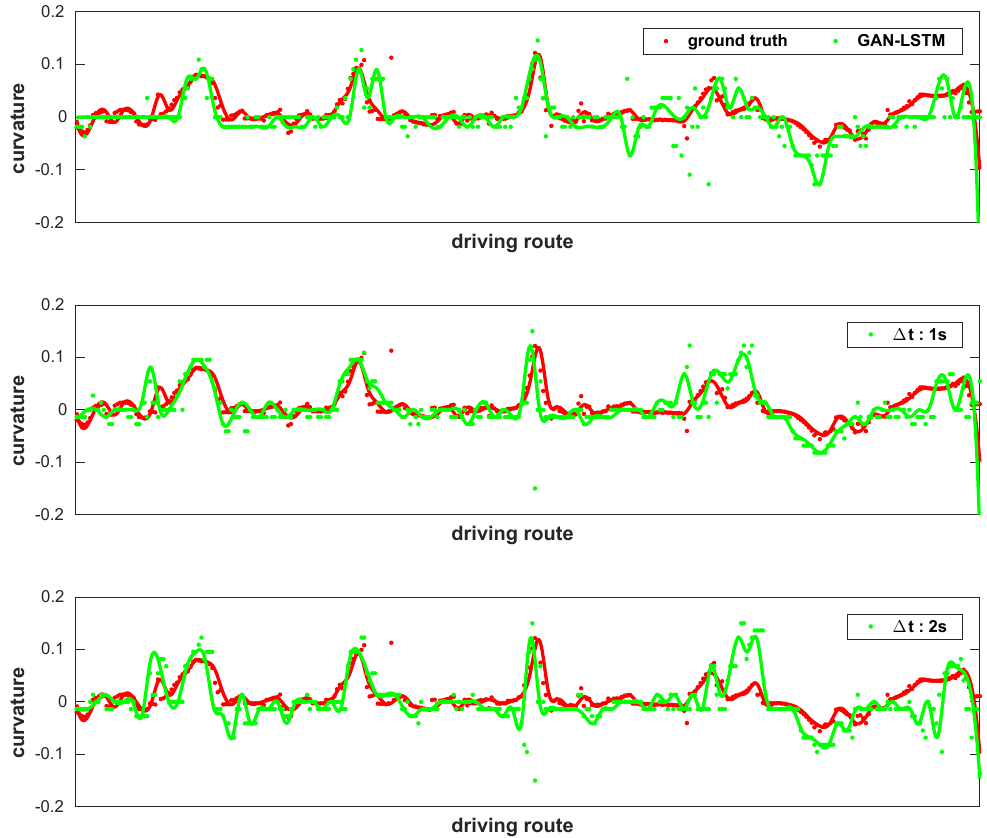}
\caption{Comparison of proposed method with human demonstration when visual prediction is delayed. From top to bottom: no time delay, 1s delay, and 2s delay. Green color indicates the model prediction and red color denotes human demonstration. The dots denoted actual data points, which are fitted to a smooth line to show the prediction tendency.}
\label{delay2}
\end{figure}

For the design of end2end approach, motion generation is correlated with environment understanding. This makes the system has a strong reliance to real-time vision prediction. However, the vision processing alone is prone to be disturbed and the GPS signal for local planned route can be lost due to occlusions. Thus, it can be a critical ability to generate valid motion when visual result is delayed. In this section, we investigate the motion generation ability when there are different level of time delays for visual prediction. The results are shown in Fig.\ref{delay1} and Fig. \ref{delay2} for end2end method and the proposed method respectively.

The figures show sequential prediction curvatures along a section of test route. The discrete dots represent actual data points, which are fitted with smooth lines to indicate variation trend. End2end approach outputs direct driving commands without intermediate knowledge retainment. Thus, during time delay, it has no choice but to keep the former motion command without human intervention. This is reflected on the extended dashed lines for motion generation in Fig. \ref{delay2}.

\begin{figure}[!h]
\centering
\includegraphics[width=\textwidth]{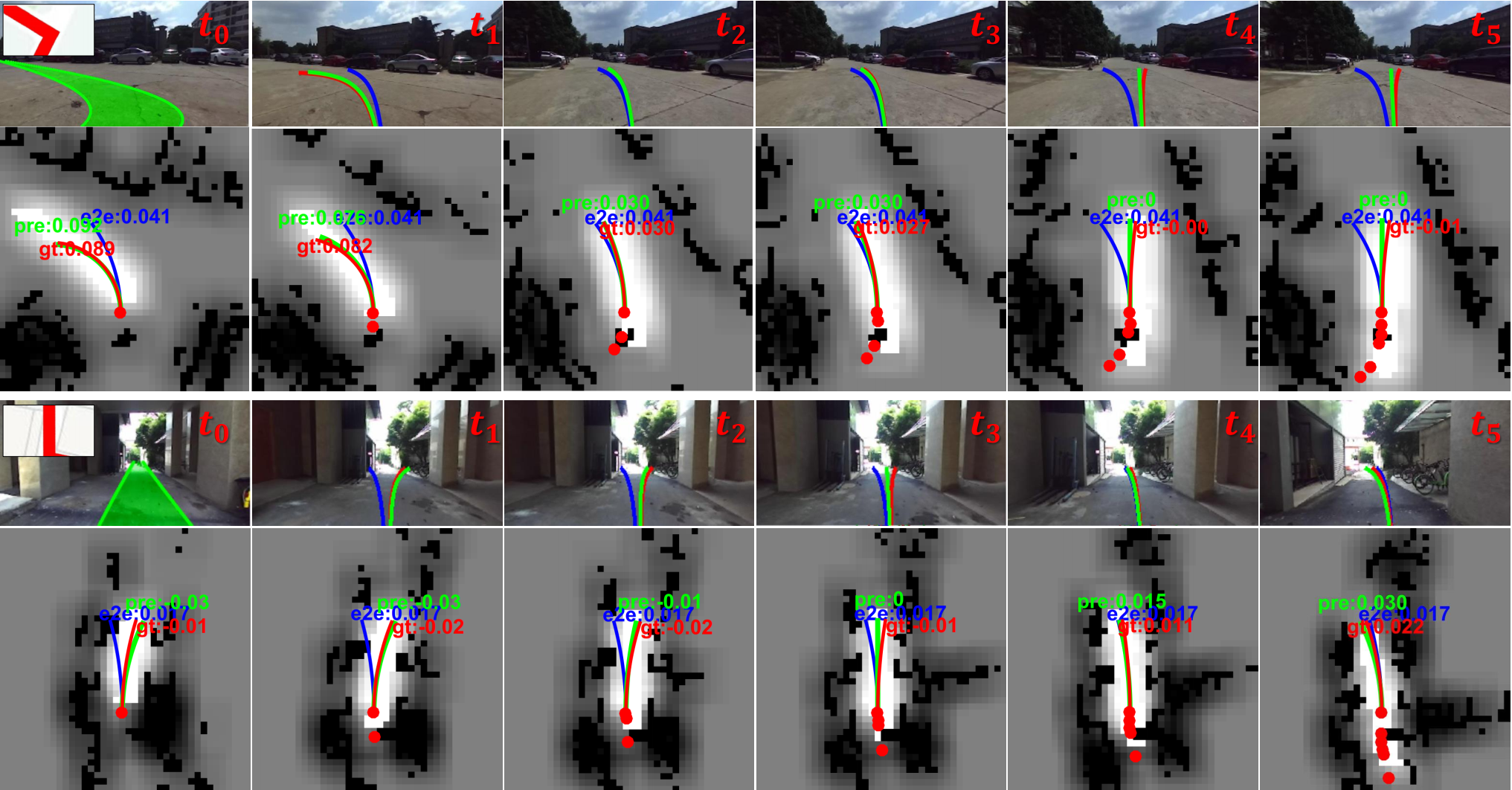}
\caption{Robustness to time delay. The time difference between consecutive frames is $\delta t=0.3s$. Green, blue, and red colors indicate motion generation results of cGAN-LSTM, end2end, and human demonstration respectively. The red points in the navigation score map indicates vechile previous pose.}
\label{timedelay}
\end{figure}

Nevertheless, for the proposed method, driving intention prediction is separated with the motion generation. The learned intention has a certain area on the ground plane, which can be used for motion reference in multiple steps. Thus, driving intention in one frame can be efficiently integrated with following obstacle perception, and be used to render a new navigation score map for motion generation. As can be inferred in Fig .\ref{delay1}, $cgan\_lstm$ model is able to generate new commands when there are temporary missing of vision predictions. This is reflected on the similar level of dispersion on the data points. Compared with the real-time visual prediction, most prediction errors for $cgan\_lstm$ appear as big curvatures. It is caused by planning circle actions for vehicle when it almost passed the retained intention area. In this case, the area around vehicle position has the biggest scores and the model tend to give the largest curvature to stay nearby. Therefore, the longest time that the proposed method can handle depends on the valid area from driving intention projection and the specific driving speed of vehicle. Some visual results for a delay of 1.8s are in our experiment shown in Fig.\ref{timedelay}

For the proposed approach, the first frame generates a visual driving intention according to local planned route, which is rendered into a navigation score map with laser perception integrated. Then for the following frames, the navigation score maps are integrated from the driving intention in the first frame and the laser perceptions in the subsequent frames. To indicate vehicle movements inside the retained driving intention area, we also plot vehicle previous poses with discrete red dots as shown in the figure. The short transformation of local coordinates can be approximately estimated by map registration of laser perception or the integration of motion command. 

In summary, the proposed method separates driving variations of visual understanding to motion generation. Compared with end2end approach, it demonstrates better prediction performance and robustness to time delay.

\section{CONCLUSIONS}\label{conclusion}
In this paper, we developed an innovative learning model for automotive driving research with an intermediate driving intention learned from image perception and publicly available route planer. The driving intention can be efficiently encoded into a navigation score map which is able to directly generate motion. In this way, the variations on motion is separated with that on visual understanding, which increases modular flexibility and reliability compared with end2end approach. The key of the system is a cGAN network inserted with LSTM unit to learn from human demonstration. The adversarial loss enables a weakly-supervised training manner, which leverages limited single-modal demonstration data to achieve generalization on multi-modal behavior in strange scenarios. Experiments indicates driving intention generation is robust to errors on vehicle global pose and shows more attributes of reliability and adaptability for consideration of real applications. Our future work will incorporate more detailed obstacle modeling modules, which further narrows the gap for real application.

\section*{ACKNOWLEDGMENT}
This work was supported in part by the National Key R\&D Program of China (2017YFB1300400), in part by the National Nature Science Foundation of China (U1609210).

\bibliography{mybibfile}

\end{document}